# Gender and Ethnicity Classification based on Palmprint and Palmar Hand Images from Uncontrolled Environment

Wojciech Michal Matkowski*, Adams Wai Kin Kong*
School of Computer Science and Engineering
Nanyang Technological University
Singapore

## Abstract

*Soft biometric attributes such as gender, ethnicity or age may provide useful information for biometrics and forensics applications. Researchers used, e.g., face, gait, iris, and hand, etc. to classify such attributes. Even though hand has been widely studied for biometric recognition, relatively less attention has been given to soft biometrics from hand. Previous studies of soft biometrics based on hand images focused on gender and well-controlled imaging environment. In this paper, the gender and ethnicity classification in uncontrolled environment are considered. Gender and ethnicity labels are collected and provided for subjects in a publicly available database, which contains hand images from the Internet. Five deep learning models are fine-tuned and evaluated in gender and ethnicity classification scenarios based on palmar 1) full hand, 2) segmented hand and 3) palmprint images. The experimental results indicate that for gender and ethnicity classification in uncontrolled environment, full and segmented hand images are more suitable than palmprint images.*

## 1. Introduction

Attributes such as gender/sex, ethnicity/race, age, weight, body mass index also known as soft biometrics have been addressed by many researchers [1]. Contrary to traditional - hard biometrics, soft biometrics are less distinctive but still may provide additional and useful information about the individual [2]. The applications of soft biometrics are, e.g., improving the performance of biometric recognition system, forensics investigation, automatic image tagging and passive demographic data collection [1].

Soft biometrics can be extracted from different biometric traits, e.g., gender [3], ethnicity [4], weight, height, body mass index [5] from face; age and gender from ear [6]; gender from gait [7], iris [8] and hand [9], etc. Surveys of soft biometrics can be found in [1] and [2]. Using soft biometrics from hand has been mostly studied based on measurements of remains, i.e., hand bones [10], [11], to infer gender for forensic, medical or archaeological applications. In biometric recognition applications such as user access control or time and attendance, hand images from controlled environments have been widely studied using, e.g., hand geometry [12], fingers [13], finger-knuckles [14], dorsal skin texture [15], vein [16] and palmprint [17], which recently also gains attention in uncontrolled environments [18]. With few exceptions, estimating soft biometrics from hand has received relatively little attention in the biometrics community.

Previous studies [9, 16, 18, 19] used hand images taken in well-controlled environment and aimed at gender classification for commercial biometric applications. However, uncontrolled environment and some forensic applications are not well investigated. In digital and multimedia forensics, child sexual assault materials (CSAM) [21], rioter and terrorist images with no obvious traits such as face, are always taken in uncontrolled and uncooperative environment. Subject's hands are visible in rioter and terrorist images during salute or waving [22], and in CSAM [23], e.g., when they touch the victim or offender. Thus, using hand image to extract soft biometrics such as gender or ethnicity, which are inherent to a subject, could give useful clues to forensic investigator and reduce a list of subjects to search.

In this study, gender and ethnicity labels are collected and provided for subjects from a publically available database and selected deep learning methods are evaluated in gender and ethnicity classification scenarios based on three types of palmar hand images taken in uncontrolled and uncooperative environment.

The rest of this paper is organized as follows. Section 2 provides related studies of soft biometrics based on hand images. Section 3 presents collected gender and ethnicity attributes for hand images. Section 4 evaluates the existing five deep learning methods in gender and ethnicity classification scenarios. Section 5 gives conclusions.

*Corresponding authors: W.M. Matkowski, email: wojciech.matkowski@ntu.edu.sg, matk0001@e.ntu.edu.sg
A.W.K. Kong, email: adamskong@ntu.edu.sg

## 2. Related work

Up until now, available biometric studies of hand[1] soft biometrics explored gender information and well-controlled imaging environment. Typically, hand-crafted as well as deep learning based approaches were proposed and evaluated in two-class classification scenarios.

Afifi [9] established a database from 190 subjects and studied recognition and gender classification using palmar and dorsal hand. Hand images were taken in well-controlled environment on a white background. For recognition and classification, the authors proposed to use two stream convolutional neural network (CNN) which takes RGB and local binary pattern (LBP) images. For the CNN, the author used pre-trained AlexNet model. Then, the author extracted features from the CNN and used SVM for gender classification achieving 94.2% and 97.3% accuracy for palmar and dorsal hand, respectively.

Xie *et al.* [15] established a database from 80 subjects and studied gender classification based on skin texture from hand dorsal images taken in well-controlled environment, achieving 98.65% accuracy. To capture skin texture, the authors proposed an imaging device which takes images under relatively high resolution of 450 dpi.

Xie *et al.* [19] used a pre-trained modified VGG network for gender classification on publically available, well-controlled multispectral (250 subjects) and contactless (312 subjects) palmprint databases, reporting 75.17% and 89.2% average accuracy, respectively.

Motivated by anthropometry and psychology, Amayeh *et al.* [20] investigated gender classification from palmar hand shape using a database from 20 male and 20 female subjects in well-controlled environment. The authors used hand-crafted features from segmented hand silhouette, and evaluated 3 different classifiers, namely, Minimum Distance, k-Nearest Neighbors (k-NN) and Linear Discriminant Analysis (LDA). The highest reported accuracy 98% was achieved by LDA. Note that the method from [20] cannot be directly applied to images considered in this study because it requires well-controlled environment and user cooperation to extract the desired features.

Matkowski *et al.* [22] established a large NTU-PI-v1 database of palmar hand images downloaded from the Internet, to study palmprint recognition in uncontrolled and uncooperative environment for forensic applications. However, soft biometric information such as gender and ethnicity is not available. Thus, in this paper, for each of 1093 subjects in NTU-PI-v1 database, which is the largest publically available palmprint image database in terms of subjects, ethnicity and gender labels are collected and provided.

## 3. Labelled hand attributes

In this paper, class labels of two attributes are provided: gender and ethnicity. Gender refers to biological sex, whereas ethnicity refers to person's physical appearance related to biological factors also known as race in the literature [1].

The NTU-PI-v1 database is the only palmprint database collected from uncontrolled environments. Thus, it is selected for this study. It contains 7881 palmar hand images of 1093 subjects and corresponding segmented hand images and hand landmarks which can be used to extract palmprint region of interest (ROI) [22]. Hand image sizes range from 30 x 30 to 1415 x 1415 pixels. Fig. 1 shows image size distribution in NTU-PI-v1. This database was initially established to study biometric recognition and contained identity labels only. However, other, e.g., soft biometric labels were not provided. In this work, for each of 1093 subjects in this database, gender and ethnicity class labels are searched and collected from the Internet and will be available online for research purpose after this work is published.

For gender, there are two labels considered: either male or female. Table 1 shows the number of male and female samples in the database. Fig. 2 shows examples of male and female hands in the database. Following the most popular and recent forensic anthropology terminology, the subjects are also classified into three different classes namely Asian, White, and Black [24]. Thus, for ethnicity, three labels are collected: 1) Asian mainly containing Chinese, Japanese, Korean and South East Asians; 2) White mainly containing people from Europe, Middle East and the Americas; 3) Black mainly containing African Americans and Africans. If a subject is a descendant of two different ethnicity persons, then he/she has two labels. E.g. if a subject's one parent is Asian and the other parent is White, then the subject is labelled as Asian and White. Totally, there are 255 images of 33 subjects with more than one ethnicity label. Note that multi-labelling has been done for data collection purpose and this multi-label [2] classification problem is not further investigated in this paper and left for future work. Table 2 shows the number of images and subjects falling into different ethnicity classes. Fig. 3 shows hand image examples of different ethnicity.

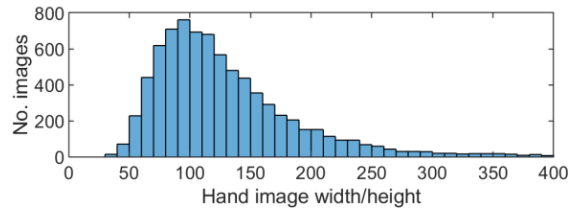

Figure 1: Distribution of hand image width/height in NTU-PI-v1 database. For better visualization, the range of horizontal axis is clipped at 400 because only ~1% of images fall beyond 400.

---

[1] Fingerprints with available features such as minutiae and ridges have also been applied to soft biometrics. Nevertheless, they are out of scope in this paper because they require high resolution image.

[2] Multi-label classification refers to a classification problem where each sample can be classified into multiple classes. In this work, in experiments, each sample can be classified into only one class.

Table 1. Distribution of gender classes in NTU-PI-v1.

| Class | No. images | No. subjects |
|---|---|---|
| Male | 4810 | 655 |
| Female | 3071 | 438 |

Table 2. Distribution of ethnicity classes in NTU-PI-v1.

| Class | No. images | No. subjects |
|---|---|---|
| White | 3894 | 588 |
| Asian | 3529 | 444 |
| Black | 679 | 94 |

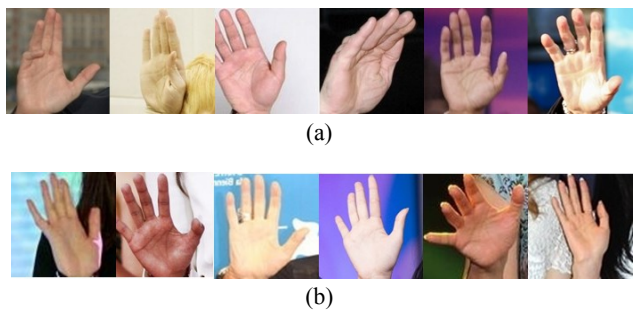

Figure 2: Examples of (a) male and (b) female hand images in NTU-PI-v1.

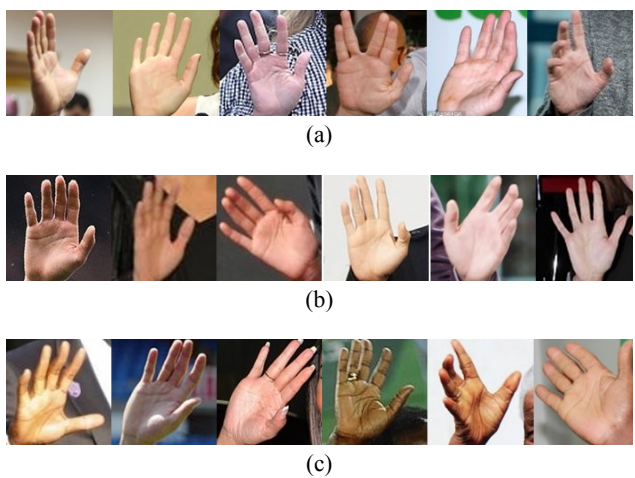

Figure 3: Examples of hand images from (a) White, (b) Asian and (c) Black subjects in NTU-PI-v1.

## 4. Experiments and results

In the experiments, gender and ethnicity classification scenarios are performed and the results are reported in Sections 4.3 and 4.4, respectively. Section 4.1 describes the experimental protocol and evaluation metrics. Section 4.2 provides five selected deep learning methods, implementation details and three types of images used in experiments.

### 4.1. Protocol and metrics

In the experiments, NTU-PI-v1 database is split into subject-disjoint training, validation and testing sets. 70%, 15% and 15% of different subjects are used for training, validation and testing, respectively. Each experiment is repeated five times using a different database split controlled by predefined random seeds, and average accuracy and correct classification rates for each class with standard deviations are reported. Accuracy, which is defined as a correct attribute classification rate is a common metric used to evaluate algorithm's performance based on soft biometrics in classification scenarios [8], [9], [3], [4], [20], on well-balanced datasets.

### 4.2. Classification algorithms and implementation details

Five state-of-the-art, pre-trained deep neural networks from PyTorch models zoo, namely AlexNet [25], DenseNet [26], ResNet-50 [27], SqueezeNet [28] and VGG-16 [29] are selected for evaluation in each experiment. Note that AlexNet and modified VGG have also been previously used in gender classification from hands in [9] and [19]. Three different types of images from NTU-PI-v1 database are considered as input: full hand image, segmented hand image, and palmprint ROI (see Fig. 4). The input images are resized to 224 x 224 pixels and normalized following the network's requirements. For augmentation, random horizontal flip, random rotation, translation, and scale are applied on-the-fly to compensate differences between hand poses in different images. In training, cross entropy loss is used and networks are fine-tuned using ADAM [30] optimizer with, learning rate 0.001 and 0.0001 for the last and all other layers, respectively. The batch size is 64 and number of epochs is 10, which always results in convergence. Networks achieving the highest accuracy on validation set are deployed for testing. The experiments are implemented in Python using PyTorch library and run on a single GPU card. Totally, there are 150 experiment runs: 5 repetitions × 5 networks × 3 types of images × 2 classification scenarios.

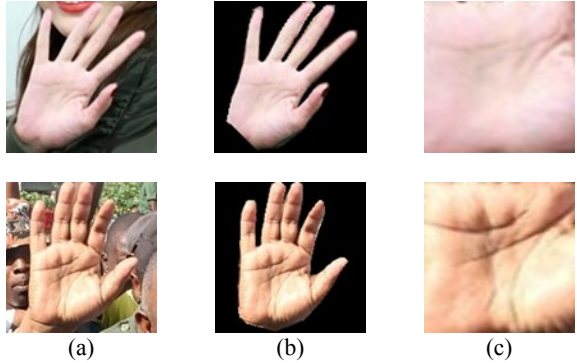

Figure 4: Three different types of images from NTU-PI-v1 database considered in experiments. (a) full hand image and the coresponding (b) and (c) segmented hand image and palmprint ROI extracted based on hand landmarks, respectively. The images in first and second row are from subjects of different gender and ethnicity.

### 4.3. Gender classification

Gender classification is performed as a two-class classification because there are male and female classes available in the labelled NTU-PI-v1 database. The experimental results for full hand image, segmented hand image and palmprint ROI (see Fig. 4) are reported in Tables 3, 4 and 5, respectively. Examples of correctly classified and misclassified full hand images are presented in Fig. 5.

Generally, DenseNet, ResNet-50 and VGG-16, achieve comparable but higher than AlexNet and SqueezeNet accuracy in all three experiments. The highest reported gender classification accuracy 88.13% is achieved by fine-tuning DenseNet using full hand images. Correct Male classification rates (CCR-Male) tend to be slightly higher than correct Female classification rates (CCR-Female), which is likely caused by some imbalance in class distribution (see Table 1). On average, CCR-Male are 0.13, 3.50 and 8.67 percentage points (p.p.) higher than CCR-Female for full hand, segmented and palmprint ROI images, respectively. Interestingly, the average accuracies for full hand images are 0.47 - 2.63 p.p. higher than for segmented hand images. One possible explanation could be that full hand images contain background which may also carry some gender related information learnt by networks, e.g., clothes, accessories, hair, body, head, etc. (see Figs. 5 and 6). Using palmprint ROI gives the lowest gender classification accuracy, e.g. DenseNet on full hand images outperforms DenseNet on palmprint ROI by 8.18 p.p.

The experimental results highlight the importance of hand geometry in gender classification from hand, which is in accordance with the previous studies [10], [20]. Note that we do not explicitly use features from hand geometry but let networks to learn features. Therefore, the experimental results indicate how much gender related information, the selected networks are able to extract from 1) palmprint ROI containing, e.g., flexion creases, wrinkles, skin color and texture; 2) segmented hand image containing 1) and fingers and hand shape; 3) full hand image containing 1) and 2) and background with, e.g., clothes, accessories, etc. Fig. 6 shows examples of gender classification results from three corresponding experiments. The only difference between these three experiments was the type of input image. It is noted in our preliminary analysis and also pointed out in Fig. 6 that in some cases, one type of image is misclassified, whereas two other corresponding images are classified correctly. It could be caused by background noise but should be investigated more in future work. For example, in Fig. 6a (second row) male hand image containing female face in background is misclassified as female. Also, misclassified female hand image in Fig. 6b (second row), contains shirt's sleeve which is more likely to be visible in male hand images.

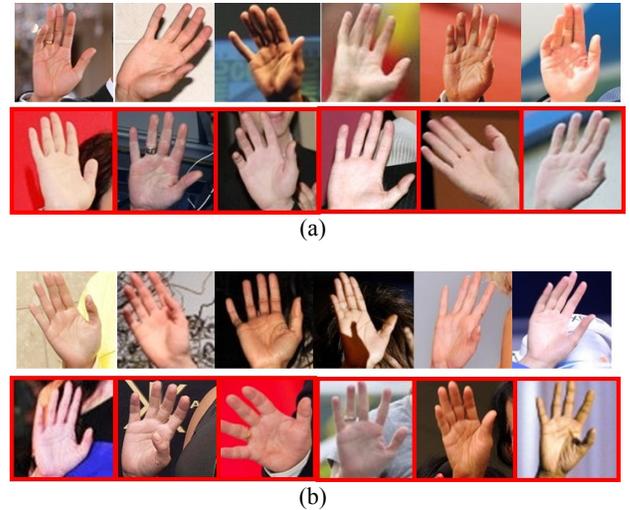

Figure 5: Examples of classification results of (a) male and (b) female full hand images. Misclassified images are in red boxes.

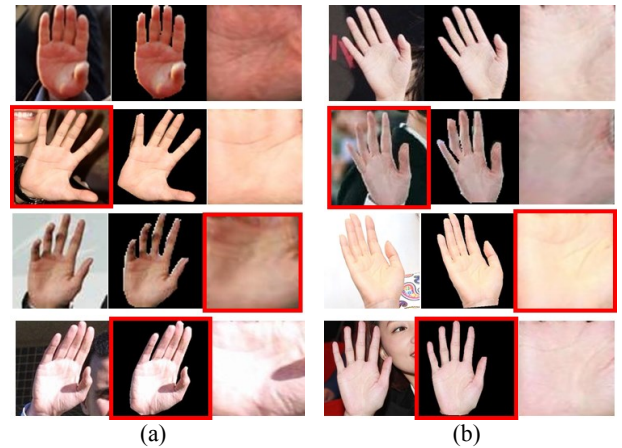

Figure 6: Examples of correctly classified and misclassified (a) male and (b) female hand images using fine-tuned DenseNet networks and three different images as input. Misclassified images are in red boxes. In each subfigure (a) and (b), full hand, segmented, and palmprint ROI images in the same row are from the same subject.

Comparing with previous works [9], [15], [19], [20] (see Section 2) of gender classification from hand, accuracy reported in this study is around 10 p.p. lower. However, the images used in this study come from uncontrolled and uncooperative environment and some of them are low quality and low resolution (see Fig. 1). Thus, lower accuracy can be expected [22], [31]. Note that in forensic applications, it is common that images are low quality and taken in uncontrolled environment.

Table 3. Gender classification accuracy and correct classification rates (CCR) of the deep neural networks using full hand images.

| Network | Accuracy (%) | CCR-Male (%) | CCR-Female (%) |
| --- | --- | --- | --- |
| AlexNet | 82.14 ± 0.81 | 83.24 ± 3.19 | 80.27 ± 3.70 |
| DenseNet | **88.13** ± 1.01 | 87.75 ± 1.36 | 88.96 ± 1.58 |
| ResNet-50 | 87.87 ± 1.45 | **89.40** ± 2.03 | 85.37 ± 2.16 |
| SqueezeNet | 83.48 ± 2.08 | 84.39 ± 2.22 | 82.39 ± 4.61 |
| VGG-16 | 86.14 ± 1.77 | 84.10 ± 3.19 | **91.22** ± 2.73 |

Table 4. Gender classification accuracy and correct classification rates (CCR) of the deep neural networks using segmented hand images.

| Network | Accuracy (%) | CCR-Male (%) | CCR-Female (%) |
| --- | --- | --- | --- |
| AlexNet | 81.66 ± 1.09 | 85.52 ± 1.75 | 75.88 ± 2.59 |
| DenseNet | **86.52** ± 0.75 | 87.21 ± 1.69 | 85.48 ± 3.78 |
| ResNet-50 | 85.23 ± 0.91 | **87.23** ± 2.16 | 82.57 ± 5.85 |
| SqueezeNet | 81.56 ± 2.04 | 82.70 ± 2.66 | 79.82 ± 4.17 |
| VGG-16 | 85.58 ± 1.09 | 85.44 ± 2.44 | **86.83** ± 4.61 |

Table 5. Gender classification accuracy and correct classification rates (CCR) of the deep neural networks using palmprint ROI.

| Network | Accuracy (%) | CCR-Male (%) | CCR-Female (%) |
| --- | --- | --- | --- |
| AlexNet | 77.30 ± 1.05 | 79.29 ± 1.36 | 73.39 ± 1.80 |
| DenseNet | **79.95** ± 0.97 | 83.33 ± 2.69 | **74.94** ± 5.20 |
| ResNet-50 | 79.04 ± 1.48 | 83.46 ± 2.08 | 72.42 ± 3.82 |
| SqueezeNet | 75.29 ± 1.35 | 78.19 ± 1.27 | 69.98 ± 2.17 |
| VGG-16 | 79.69 ± 0.89 | **83.49** ± 2.14 | 73.67 ± 1.87 |

### 4.4. Ethnicity classification

Ethnicity classification is performed as a three-class classification, where each subject is being classified into either Asian, White or Black class (see Section 3). A number of Black subjects in NTU-PI-v1 is relatively low (see Table 2). Thus, in training and validation sets, this class is oversampled. If a subject has multiple labels (see Section 3), then he/she is relabeled for the experimental purpose and merged with the shared majority class. The experimental results are summarized in Tables 6, 7 and 8 for full hand, segmented and palmprint ROI images (see Fig. 4), respectively. Fig. 7 shows examples of classified full hand images.

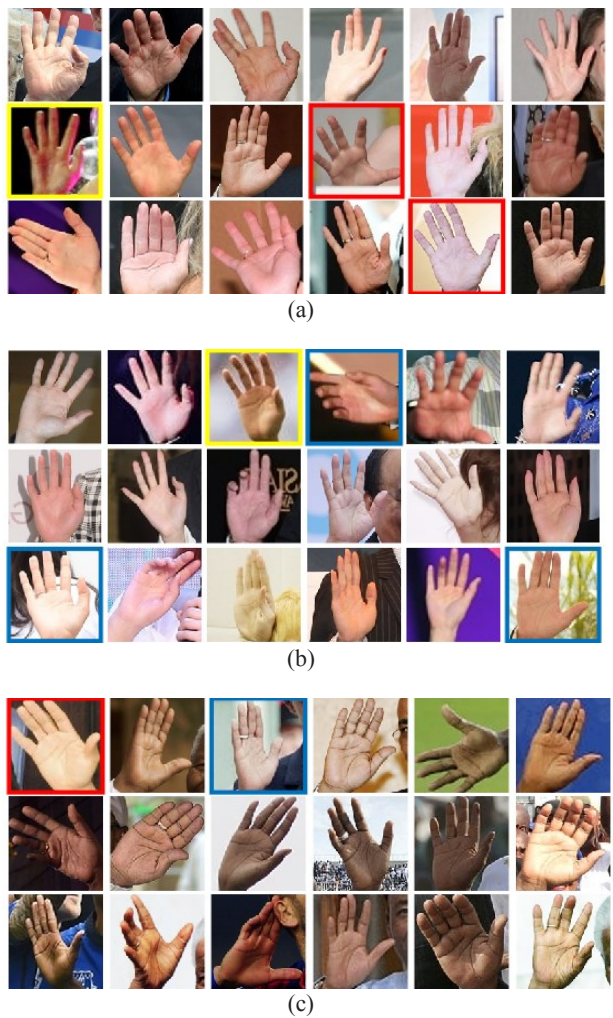

Figure 7: Examples of correctly classified and missclassified full hand images from (a) White, (b) Asian and (c) Black subjects. Red, blue and yellow boxes indicate missclasifications as Asian, White and Black, respectively.

DenseNet achieves the highest ethnicity classification accuracy of 81.09% on the segmented hand images outperforming the second best ResNet-50 by 1.79 p.p. CCR-White is on average 3.86, 4.28 and 2.66 p.p. lower than CCR-Asian, on full hand, segmented and palmprint ROI, respectively. It could be related to the fact that the White class considered in this work contains populations from various continents (see Section 3) with relatively high diversity. It should be noted that the reported average CRC-Black come from a relatively small sample size of testing set (around 70 images) comparing to CRC-Asian (around 500 images) and CRC-White (around 500 images). In addition, CRC-Black standard deviations are 9.90 – 18.81 p.p. Thus the CRC-Black results should be treated with less significance because they are highly affected by random split sampling of a relatively small number of subjects.

Table 6. Ethnicity classification accuracy and correct classification rates (CCR) using full hand image.

| Network | Accuracy (%) | CCR-White (%) | CCR-Asian (%) | CCR-Black (%) |
|---|---|---|---|---|
| AlexNet | 70.58 ± 2.11 | 72.42 ± 7.28 | 68.66 ± 4.02 | 71.47 ± 11.77 |
| DenseNet | **78.97** ± 2.09 | **78.88** ± 1.35 | 79.79 ± 3.54 | 74.97 ± 15.69 |
| ResNet-50 | 77.86 ± 1.88 | 76.14 ± 3.87 | 81.01 ± 5.15 | 71.12 ± 16.79 |
| SqueezeNet | 68.86 ± 3.15 | 65.47 ± 9.12 | 71.82 ± 6.35 | 75.28 ± 15.50 |
| VGG-16 | 75.79 ± 1.35 | 70.53 ± 6.33 | **81.36** ± 5.41 | **75.45** ± 12.49 |

Table 7. Ethnicity classification accuracy and correct classification rates (CCR) using segmented hand image.

| Network | Accuracy (%) | CCR-White (%) | CCR-Asian (%) | CCR-Black (%) |
|---|---|---|---|---|
| AlexNet | 74.85 ± 1.95 | 73.17 ± 2.63 | 76.91 ± 1.22 | 74.19 ± 15.14 |
| DenseNet | **81.09** ± 1.88 | 78.36 ± 5.06 | **85.34** ± 4.03 | 74.33 ± 12.38 |
| ResNet-50 | 79.33 ± 2.64 | **80.99** ± 2.86 | 79.33 ± 4.97 | 69.98 ± 14.32 |
| SqueezeNet | 73.47 ± 4.09 | 68.27 ±10.78 | 78.74 ± 6.11 | **77.41** ± 9.90 |
| VGG-16 | 77.74 ± 4.01 | 77.67 ± 5.87 | 79.54 ± 7.37 | 68.49 ± 13.78 |

Table 8. Ethnicity classification accuracy and correct classification rates (CCR) using palmprint ROI.

| Network | Accuracy (%) | CCR-White (%) | CCR-Asian (%) | CCR-Black (%) |
|---|---|---|---|---|
| AlexNet | 67.41 ± 2.53 | 67.34 ± 9.15 | 67.78 ± 4.66 | 67.06 ± 17.19 |
| DenseNet | 71.69 ± 1.65 | **70.80** ± 5.34 | 73.72 ± 6.38 | 67.86 ± 18.81 |
| ResNet-50 | **73.14** ±2.16 | 69.31 ± 4.95 | **77.97** ± 4.96 | 69.34 ± 14.38 |
| SqueezeNet | 66.73 ± 0.69 | 67.70 ± 6.99 | 64.26 ± 10.79 | **75.94** ±17.39 |
| VGG-16 | 70.22 ± 2.89 | 67.85 ± 5.72 | 72.60 ± 2.83 | 71.14 ± 14.63 |

As with the experimental results from Section 4.3, using palmprint ROI, provides less discriminative information comparing with full and segmented hand images. The highest reported accuracy for full hand, segmented and palmprint ROI image are 78.97%, 81.09% and 73.14%, respectively. These results also indicate that hand shape is a contributing trait in ethnicity classification based on considered hand images. Moreover, using segmented hand image results in higher classification accuracy than full hand image, where background is visible. This is in contrast to gender classification results, where background can also provide some useful information. These experimental results suggest that differences in hand image backgrounds containing clothes, part of body or hair are higher between gender classes, rather than ethnicity classes.

## 5. Conclusions

In this preliminary work, gender and ethnicity classification based on palmar hand images taken in uncontrolled environment are investigated. Gender and ethnicity labels for all the subjects in NTU-PI-v1 database are provided. Five selected deep neural networks are fine-tuned and evaluated in gender and ethnicity classification scenarios on full hand, segmented, and palmprint ROI images. The experimental results indicate that full hand and segmented hand images provide more discriminative information than palmprint ROI image and thus higher accuracy in both gender and ethnicity classification experiments. The highest reported accuracy for gender and ethnicity classification are 88.13% and 81.09%, respectively. In future, joint gender and ethnicity classification as well as fusion of different hand image types could be considered. We also plan to use gender and ethnicity information to improve the hand and palmprint recognition performance.

## Acknowledgment

This work is partially supported by the Ministry of Education, Singapore through Academic Research Fund Tier 1, RG21/19 (2019-T1-001-102). The authors would like to thank Soohyun Park from MSE, NTU for help with establishing ethnicity labels, including multiple discussions and searching the Internet.